\documentclass[10pt,twocolumn,letterpaper]{article}

\usepackage{wacv}              

\usepackage{graphicx}
\usepackage{amsmath}
\usepackage{amssymb}
\usepackage{booktabs}
\usepackage{svg}
\usepackage[accsupp]{axessibility}  

\usepackage[pagebackref,breaklinks,colorlinks]{hyperref}

\usepackage{epsfig}
\usepackage{tabularx}
\usepackage{multirow,bigdelim}
\usepackage{enumitem}
\usepackage{bm}
\usepackage[symbol]{footmisc}
\usepackage[font=footnotesize,labelfont=bf]{caption}
\usepackage{pifont}
\newcommand{\xmark}{\ding{55}}%

\newcommand{\datasetSampleHist}{
\begin{figure*}[t]
  \centering
  \includegraphics[width=0.95\linewidth]{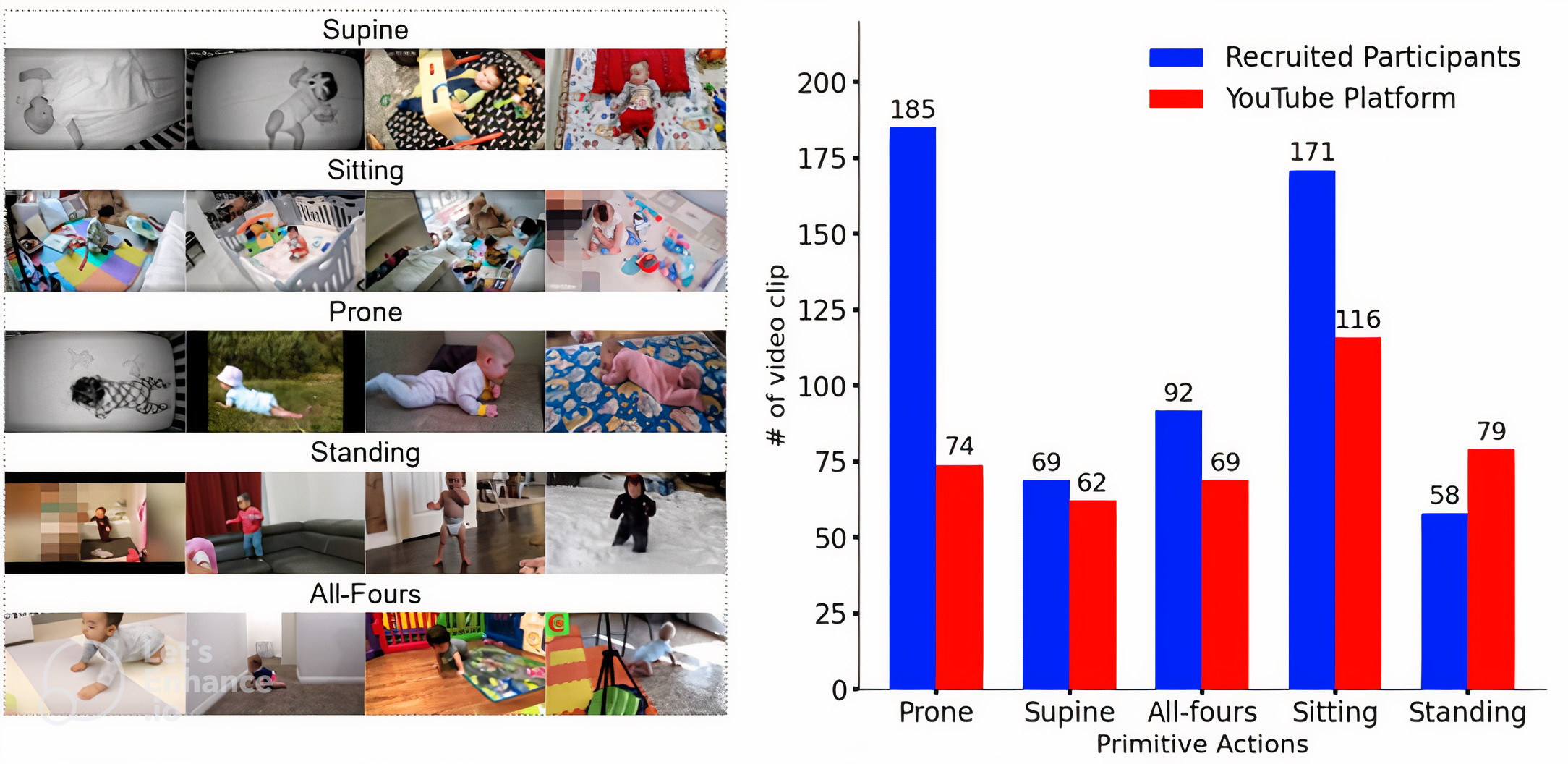}
  \caption{Some snapshots from the InfActPrimitive dataset are displayed on the left side. Each row corresponds to one of the five infant primitive action classes of the dataset. On the right side, the frequency of each action class is depicted, collected from both the YouTube platform and our recruited participants through an IRB-approved experiment.}
  \label{fig:examples}
\end{figure*}
}

\newcommand{\framework}{
\begin{figure*}[t]
  \centering
  \includegraphics[width=0.95\linewidth]{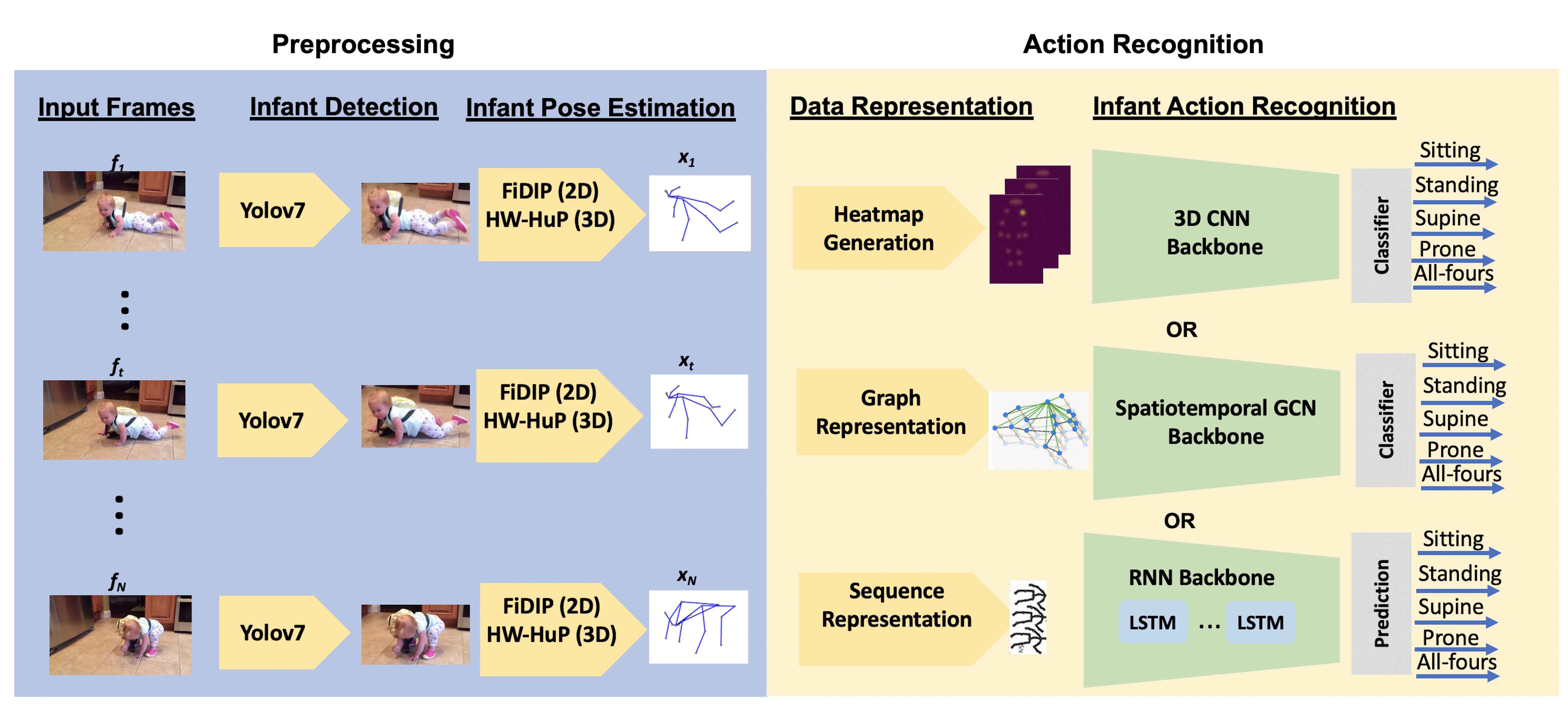}
  \caption{Schematic of the overall infant action recognition pipeline, encompassing infant-specific preprocessing and the action recognition phase. The infant is initially detected in raw frames using YOLOv7 \cite{Yolo7} and subsequently serves as input for both 2D and 3D pose estimation facilitated by FiDIP \cite{huang2021invariant} and HW-HuP-Infant \cite{liu2021heuristic} algorithms, respectively. The resulting pose information can be further processed into heatmaps, serving as input for CNN-based models, or represented as graphs or sequences for graph- and RNN-based models to predict infant actions.}
  \label{fig:frame}
\end{figure*}
}

\newcommand{\skelayout}{
\begin{figure}[t]
\centering
\includegraphics[width=\linewidth]{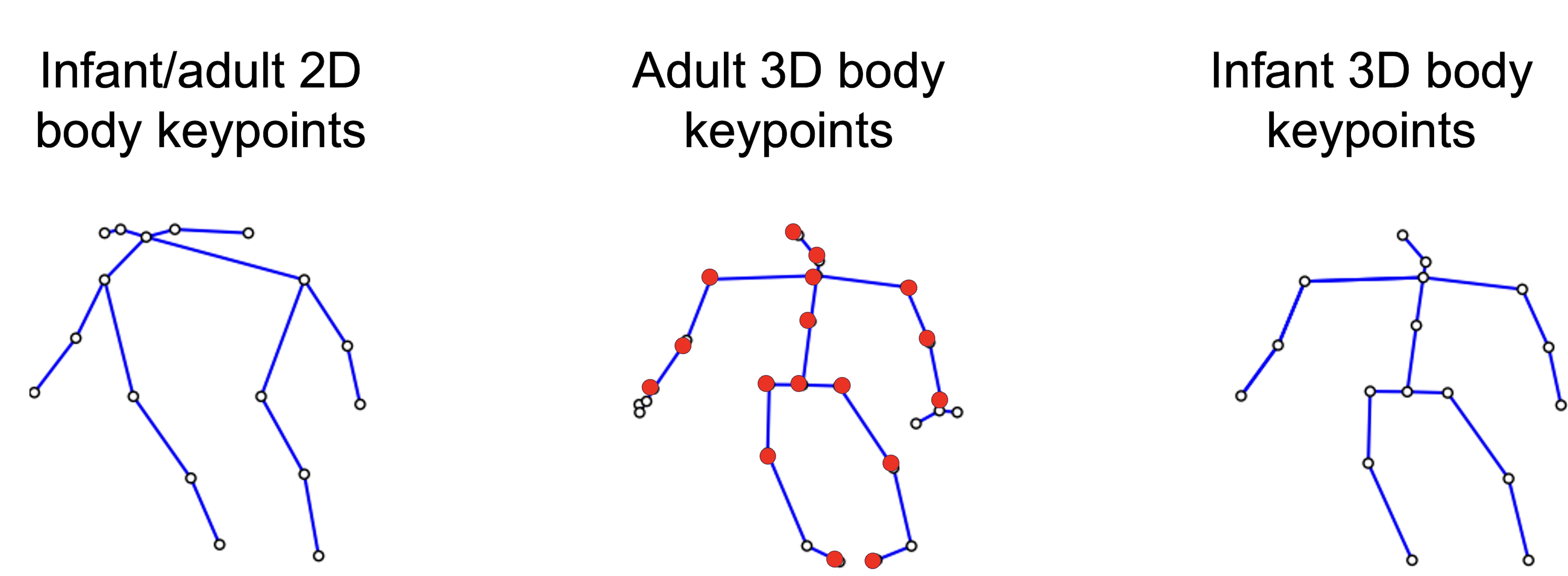}
\caption{Visualization of three distinct skeleton layouts employed in skeleton-based action recognition datasets. The adult skeleton data adheres to the NTU RGB+D layout, while the 3D version of InfActPrimitive adopts the Human3.6M layout. Action recognition models utilize the common keypoints shared between these layouts, highlighted in red. Additionally, both the 2D versions of adult and infant skeleton data conform to the COCO layout.}
\label{fig:layout}
\end{figure}
}

\newcommand{\confmatrixes}{
\begin{figure*}[t]
  \centering
  \includegraphics[width=0.95\linewidth]{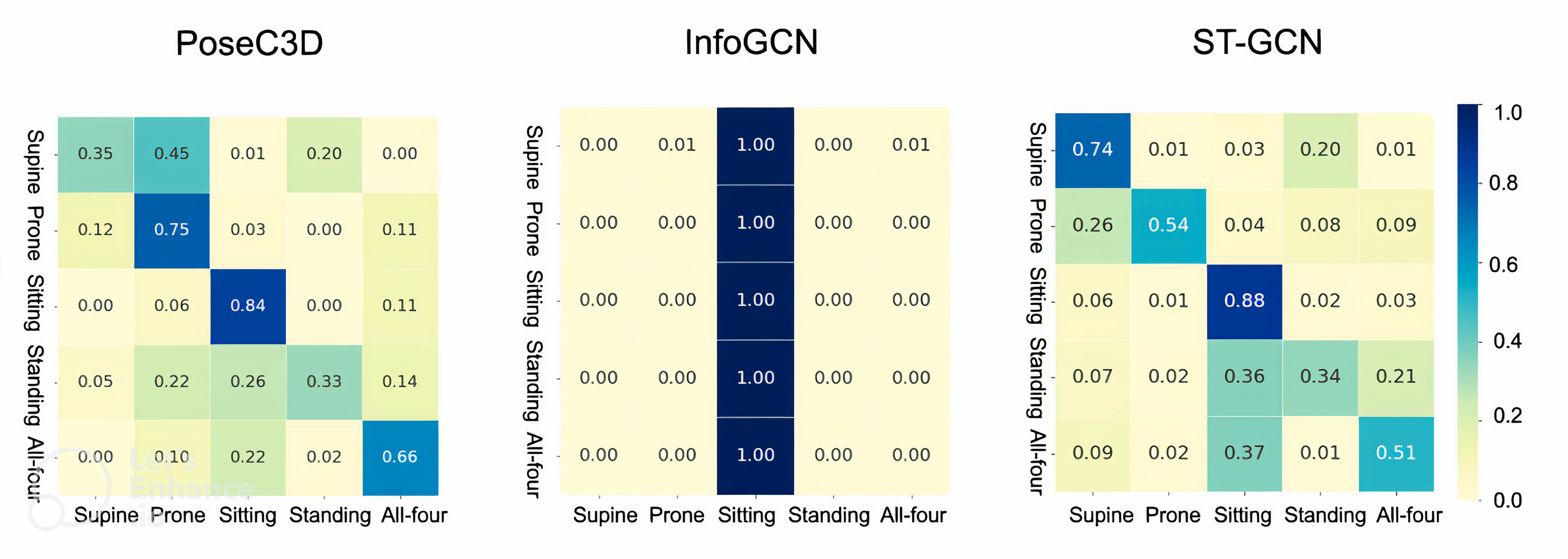}
  \caption{The classification results of three models, along with their respective confusion matrices, are displayed. As shown, InfoGCN faces challenges in achieving clear distinctions between classes, whereas the other models demonstrate varying degrees of proficiency in classifying different primitive categories.}
  \label{fig:confs}
\end{figure*}
}

\newcommand{\tsne}{
\begin{figure*}[t]
\centering
\includegraphics[width=\linewidth]{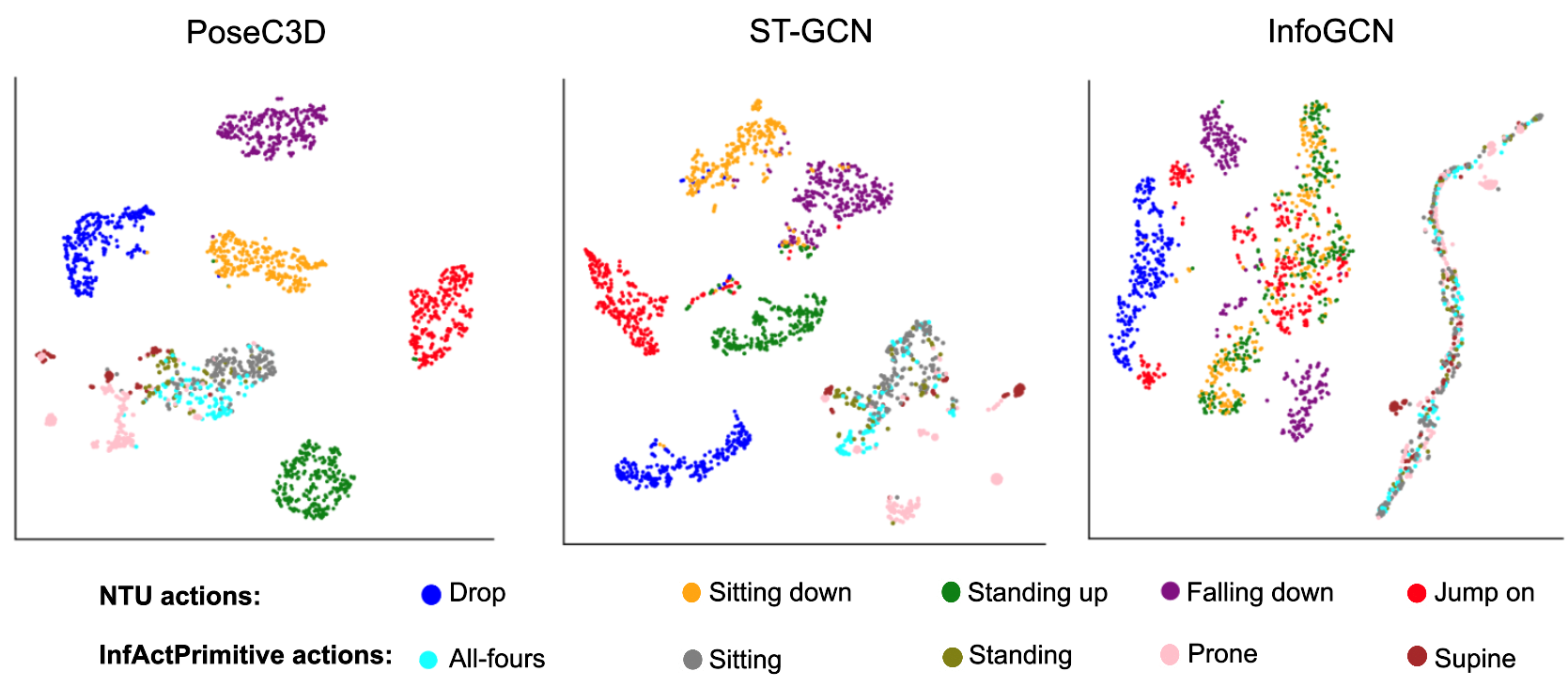}
\caption{2D latent projections generated through t-SNE for validation samples from both the NTU RGB+D and InfActPrimitive datasets. The results, presented from left to right, demonstrate the projection of the latent variables produced by PoseC3D, InfoGCN, and ST-GCN. While these methods effectively capture patterns in adult actions within the NTU RGB+D dataset, they struggle to distinguish between infant actions in the InfActPrimitive dataset.}
\label{fig:tsne}
\end{figure*}
}

\newcommand{\posresults}{
\begin{table*}[t]
    \centering
    \footnotesize
    \caption{Results of 2D/3D skeleton-based action recognition models using our proposed pipeline on both adult (NTU RGB+D) and infant (InfActPrimitive) dataset. FT denotes that the model was pre-trained on  NTU RGB+D during the transfer learning experiments. PoseC3D achieves the best performance on 2D data in both adult and infant datasets. PoseC3D only supports 2D data, and the results in 3D space are marked with \xmark.The DeepLSTM model also resulted in very unsatisfactory performance when applied to 3D skeleton data, which we denoted with \xmark}
    \begin{tabular}{l|ccc|ccc}
    \toprule\toprule
     & \multicolumn{3}{c|}{Based on 2D Pose} & \multicolumn{3}{c}{Based on 3D Pose} \\
    \midrule
    Action Model &  NTU RGB+D &  InfActPrimitive & InfActPrimitive (+FT) & NTU RGB+D &  InfActPrimitive & InfActPrimitive (+FT)\\
    \midrule
    DeepLSTM \cite{shahroudy2016ntu} & 87.0 &  24.3 &  17.2  &  \xmark & \xmark & \xmark\\
    ST-GCN \cite{stgcn}   & 81.5 &  64.0 &  66.9  & 82.5 &  \textbf{67.1} &  \textbf{69.7}\\
    InfoGCN  \cite{chi2022infogcn} &  91.0 & 29.7 &  29.7  & \textbf{85.0} &  29.7 &  29.7\\
    PoseC3D  \cite{posec3d} &  \textbf{94.1} & \textbf{66.9} &  \textbf{69.7}  & \xmark & \xmark & \xmark\\
    \bottomrule
    \end{tabular}
    \label{tab:allResults}
\end{table*}
}

\newcommand{\crossvalresults}{
\begin{table}[]
    \centering
    \caption{Infant action recognition results with inter-class data diversity using PoseC3D \cite{posec3d}. InfActPrimitive training set is partitioned into five folds, with one fold reserved for validation while the remaining folds were used to train the model. The last row of the table presents the mean and variance computed across all folds.}
    \begin{tabular}{r|ccc}
    \toprule\toprule
     Held-out fold & Train & Validation & Test \\
    \midrule
    Fold 1 & $\bm{93.7}$ & $83.7$ & $\bm{64.3}$ \\
    Fold 2 & $87.5$ & $\bm{91.2}$ & $61.2$ \\
    Fold 3 & $\bm{93.7}$ & $83.0$ & $56.3$ \\
    Fold 4 & $\bm{93.7}$ & $78.7$ & $60.8$ \\
    Fold 5 & $\bm{93.7}$ & $85.0$ & $50.6$ \\
      \midrule
    \textbf{Average} & 92.50$\pm$ 6.2 & 84.3$\pm$ 16.3 & 58.6$\pm$22.7\\
   
    \bottomrule
    \end{tabular}
    \label{tab:cross_val_results}
\end{table}

}

\usepackage[capitalize]{cleveref}
\crefname{section}{Sec.}{Secs.}
\Crefname{section}{Section}{Sections}
\Crefname{table}{Table}{Tables}
\crefname{table}{Tab.}{Tabs.}

\def\wacvPaperID{4} 
\def\confName{WACV}
\def\confYear{2024}

\begin{document}


\renewcommand*{\thefootnote}{\fnsymbol{footnote}}

\title{Challenges in Video-Based Infant Action Recognition:\\ A Critical Examination of the State of the Art}
\author{Elaheh Hatamimajoumerd 
$^{\dagger,1,2}$, Pooria Daneshvar Kakhaki $^{\dagger, 1}$, Xiaofei Huang$^1$, Lingfei Luan$^3$,\\ Somaieh Amraee$^{1,2}$, Sarah Ostadabbas$^{1*}$ \\
$^1$Department of Electrical \& Computer Engineering,Northeastern University, MA, USA\\
$^2$ Roux Institute, Northeastern University, ME, USA\\
$^3$ University of Minnesota, MN,  USA\\
{$^*$Corresponding author's email: \tt\small Ostadabbas@ece.neu.edu}
}
\maketitle

\def\thefootnote{$\dagger$}\footnotetext{These authors contributed equally to this work.}

\begin{abstract}
Automated human action recognition, a burgeoning field within computer vision, boasts diverse applications spanning surveillance, security, human-computer interaction, tele-health, and sports analysis. Precise action recognition in infants serves a multitude of pivotal purposes, encompassing safety monitoring, developmental milestone tracking, early intervention for developmental delays, fostering parent-infant bonds, advancing computer-aided diagnostics, and contributing to the scientific comprehension of child development. This paper delves into the intricacies of infant action recognition, a domain that has remained relatively uncharted despite the accomplishments in adult action recognition. In this study, we introduce a groundbreaking dataset called ``InfActPrimitive'', encompassing five significant infant milestone action categories, and we incorporate specialized preprocessing for infant data. We conducted an extensive comparative analysis employing cutting-edge skeleton-based action recognition models using this dataset. Our findings reveal that, although the PoseC3D model achieves the highest accuracy at approximately 71\%, the remaining models struggle to accurately capture the dynamics of infant actions. This highlights a substantial knowledge gap between infant and adult action recognition domains and the urgent need for data-efficient pipeline models\footnote{The code and our data are publicly available at \href{https://github.com/ostadabbas/Video-Based-Infant-Action-Recognition}{https://github.com/ostadabbas/Video-Based-Infant-Action-Recognition}.}.
\end{abstract}

\section{Introduction}
\label{sec:intro}
Automated human action recognition is a rapidly evolving field within computer vision, finding wide-ranging applications in areas such as surveillance    , security \cite{tripathi2018suspicious}, human-computer interaction \cite{jaimes2007multimodal}, tele-health \cite{rezaei2019target}, and sports analysis \cite{wu2022survey}. In healthcare, especially concerning infants and young children, the capability to automatically detect and interpret their actions holds paramount importance. Precise action recognition in infants serves multiple vital purposes, including ensuring their safety, tracking developmental milestones, facilitating early intervention for developmental delays, enhancing parent-infant bonding, advancing computer-aided diagnostic technologies, and contributing to the scientific understanding of child development.

The notion of action in the research literature exhibits significant variability and remains a subject of ongoing investigation \cite{moeslund2006survey}. In this paper, we focus on recognizing infants' fundamental motor primitive actions, encompassing five posture-based actions (sitting, standing, supine, prone, and all-fours) as defined by the Alberta infant motor scale (AIMS) \cite{fuentefria2017motor}. These actions correspond to significant developmental milestones achieved by infants in their first year of life.


To facilitate the accurate recognition of these actions, we employ skeleton-based models, which are notable for their resilience against external factors like background or lighting variations. In comparison to RGB-based models, these skeleton-based models offer superior efficiency. Given their ability to compactly represent video data using skeletal information, these models prove to be especially useful in situations where labeled data is scarce. Therefore, their employment enables a more efficient recognition of the aforementioned hierarchy of infant actions, even with ``small data'' \cite{liu2019action}.
\datasetSampleHist

While state-of-the-art skeleton-based human action recognition and graphical convolution network (GCN) models \cite{zhang2019graph,kipf2016semi} have achieved impressive performance, they are primarily focused on the adult domain and relied heavily on large, high-quality labeled datasets. However, there exists a significant domain gap between the adult and infant action data due to differences in body shape, poses, range of actions, and motor primitives. Additionally, even for the same action, there are discernible differences in how it is performed between infants and adults. For example, sitting for adults often involves the use of chairs or elevated surfaces, providing stability and support, while infants typically sit on the floor, relying on their developing core strength and balance, resulting in different skeleton representations. Furthermore, adult action datasets like ``NTU RGB+D" \cite{shahroudy2016ntu}' and ``N-UCLA" \cite{wang2014cross} primarily include actions such as walking, drinking, and waving, which do not involve significant changes in posture. In contrast, infant actions like rolling, crawling, and transitioning between sitting and standing require distinct postural transitions. This domain gap poses significant challenges and hampers the current models' ability to accurately capture the complex dynamics of infant actions.

This paper contributes to the field of infant action recognition by highlighting the challenges specific to this domain, which has been largely unexplored despite the successes in adult action recognition. The limitations in available infant data necessitate the identification of new action categories that cannot be learned from existing datasets. To address this issue, the paper's focus is on adapting action recognition models trained on adult data for use on infant action data, considering the adult-to-infant shift, and employing data-efficient methods. 


In summary, this paper introduces several significant contributions:
\begin{itemize}
\item A novel dataset called infant action (InfActPrimitive) specifically designed for studying infant action recognition. \autoref{fig:examples} shows some snapshots of InfActPrimitive. This dataset includes five motor primitive infant milestones as basic actions.

\item Baseline experiments conducted on the InfActPrimitive dataset using state-of-the-art skeleton-based action recognition models. These experiments provide a benchmark for evaluating the performance of infant action recognition algorithms.

\item Insight into the challenges of adapting action recognition models from adult data to infant data. The paper discusses the domain adaptation challenges and their practical implications for infant motor developmental monitoring, as well as general infant health and safety.
\end{itemize}

Overall, these contributions enhance our understanding of infant action recognition and provide valuable resources for further research in this domain.



\section{Related Work}
\label{sec:related}
The existing literature on vision-based human action recognition can be classified into different categories based on the type of input data, applications, model architecture, and techniques employed. This paper focuses on reviewing studies conducted specifically on skeleton data (i.e. 2D or 3D body poses) in human action recognition. Additionally, it discusses the vision-based approaches that have been applied to the limited available infant data.

\textbf{Recurrent neural network structures} methods, such as long short term memory (LSTM) and gated recurrent unit (GRU), treat the skeleton sequences as sequential vectors, focusing primarily on capturing temporal information. However, they often overlook the spatial information present in the skeletons \cite{li2017skeleton}. Shahroody et al. \cite{shahroudy2016ntu} introduced a part-aware LSTM model that utilizes separate stacked LSTMs for processing different groups of body joints, with the final output obtained through a dense layer combination, enhancing action recognition by capturing spatiotemporal patterns. \cite{liu2017skeleton} proposed the global context-aware attention LSTM (GCA-LSTM) that incorporates a recurrent attention mechanism that selectively emphasizes the informative joints within each frame.

\textbf{Graph convolutional network (GCN)} has emerged as a prominent method for skeleton-based action recognition. It enables the efficient representation of spatiotemporal skeleton data by encapsulating the intricate nature of an action into a sequence of interconnected graphs. Spatial temporal graph convolution network (ST-GCN) introduced inter-inframe edges, connecting corresponding joints across consecutive frames. This approach enhances the modeling of inter-frame relationships and improves the understanding of temporal dynamics within the skeletal data. InfoGCN \cite{chi2022infogcn} combines a learning objective and an encoding method using attention-based graph convolution that captures discriminative information of human actions.

\textbf{3D convolutional networks} capture the spatio-temporal information in skeleton sequences using image-based representations. Wang et al. \cite{wang2016action} encoded joint trajectories into texture images using HSV space, but the model performance suffered from trajectory overlapping and the loss of past temporal information. Li et al. \cite{li2017joint} addressed this issue by encoding pair-wise distances of skeleton joints into texture images and representing temporal information through color variations. However, their model encountered difficulties in distinguishing actions with similar distances.

\textbf{Available datasets for human action recognition} are mainly incorporate RGB videos with 2D/3D skeletal pose annotations. The majority of the aforementioned studies employed large labeled skeleton-based datasets, such as NTU RGB+D \cite{shahroudy2016ntu}, which consisted of over 56 thousand sequences and 4 million frames, encompassing 60 different action classes. The Northwestern-UCLA (N-UCLA) \cite{wang2014cross} is another widely used skeleton based  dataset consists of 1494 video clips featuring 10 volunteers, captured using 3 Kinect cameras from multiple angles to obtain 3D skeletons with 20 joints, encompassing a total of 10 action categories.


\textbf{Infant-specific computer vision studies} have been relatively scarce while there have been notable advancements in computer vision within the adult domain. The majority of these studies have been primarily focused on infant images for tasks such as pose estimation \cite{huang2021invariant,9415088}, facial landmarks detection \cite{wan2022infanface,zhu2023nns}, posture classification \cite{huang2022infantposture,huang2022symmetry}, and 3D synthetic data generation \cite{liu2022heuristic}. \cite{onal2023infant} finetuned VGG-16 pretrained with adult faces for infant facial action unit recognition. They applied their methods to the CLOCK \cite{hammal2017automatic} and MIAMI \cite{chen2021person} datasets, which were specifically designed to investigate neurodevelopmental and phenotypic outcomes in infants with craniofacial microsomia and assess the facial actions of 4-month-old infants in response to their parents, respectively. Zhu et al. \cite{zhu2023nns} proposed a CNN-based pipeline to detect and temporally segment the non-nutritive sucking pattern using nighttime in-crib baby monitor footage. \cite{9515507} introduced BabyNet that uses a ResNet model followed by an LSTM to capture the spatial and temporal connection of annotated bounding boxes to interpret the onset and offset of reaching and to detect a complete reaching action. However, the focus of these studies has predominantly been on a limited set of facial actions or the detection of specific actions, thereby neglecting actions that involve diverse poses and postures. Huang et al. \cite{huang2023posture} addressed this issue by creating a small dataset containing a diverse range of infant actions and few samples for each action. The authors developed a posture classification model that was applied on every frame of an input video to extract the posture probability signal. Subsequently, a bi-directional LSTM is employed to segment the signal and estimate posture transitions and the action associated with that transition. Despite presenting a challenging dataset, their action recognition pipeline is not an end-to-end approach.

In this paper, we enhance the existing dataset initially employed in Huang et al.'s study \cite{huang2023posture} to create a more robust dataset. This expansion involves classifying actions into specific simple primitive motor actions, including "sitting," "standing," "prone," "supine," and "all-fours." Additionally, we collected additional video clips of infants in their natural environment, encompassing both daytime play and nighttime rest, in various settings such as playtime and crib environments. Finally, we tackle the intricate task of infant action recognition through a comprehensive end-to-end approach, with a specific focus on the challenges associated with adapting action recognition models from the adult domain to the unique infant domain.

\framework
\section{Methods}
\label{sec:methods}
The goal of a human action recognition framework is to assign labels to the actions present in a given video. In the infant domain, our focus is the most common actions, related to infant motor development milestones. This section introduces our dataset and pipeline for modeling infant skeleton sequences, aiming to create distinct representations for infant action recognition. We begin by introducing the InfActPrimitive dataset, which serves as the foundation for training and evaluating our pipeline. Subsequently, we delve into the details of the pipeline, which encompasses the entire process from receiving video frames as input to predicting infant action.


%

\subsection{InfActPrimitive Dataset}
\label{sec:infact}
We present a new dataset called InfActPrimitive as a benchmark to evaluate infant action recognition models. Videos in InfActPrimitive are provided from two sources. (1) Videos submitted by recruited participants: We collected infant videos using a baby monitor from their home and in an unscripted manner. The experiment was approved by the Committee on the Use of Humans as Experimental Subjects of Northeastern university (IRB number:22-11-32). Participants provided informed written consent before the experiment and were compensated for their time. (2) Videos gathered from public video-sharing platforms. This portion of video clips in our dataset has been adapted from \cite{huang2023posture}, which was acquired by performing searches for public videos on the YouTube platform. InfActPrimitive contains $814$ infant action videos of five basic motor primitives representing specific postures such as sitting, standing, prone, supine, and all four. The start and end time of every motor primitive is meticulously annotated in this dataset. The InfActPrimitive, with its motor primitives defined by the Alberta Infant Motor Scale (AIMS) as significant milestones, is ideal for developing and testing models for infant action recognition, milestone tracking, and detection of complex actions. \autoref{fig:examples} shows the screenshots from various videos within the InfActPrimitive dataset, illustrating the diversity of pose, posture, and action among the samples. The diverse range of infant ages and a wide variety of movements and postures within the InfActPrimitive dataset pose significant challenges for action recognition tasks. The right side of the panel in \autoref{fig:examples} shows the statistical analysis of InfActPrimitive for each sources of data separately. 

\subsection{Infant Action Recognition Pipeline}
\label{sec:pipeline}
Infant specific prepossessing, skeleton data prediction, and action recognition are the key components of our pipeline, as shown in \autoref{fig:frame}. To achieve this, input frames are processed through the pipeline's components, enabling infant-specific skeleton data generation and alignment as input to the different state-of-the-art action recognition models.

\textbf{Preprocessing--} Input video $V$ is represented as sequence of $T$ frames, $V = \left(f^1,\ldots,f^t,\ldots,f^T\right)$.  We customized the YOLOv7 \cite{Yolo7} to locate the bounding box around the infants at every frame as a region of interest. We then extracted either a 2D or 3D infant skeleton pose prediction $x^t \in \mathbb{R}^{J\times D}$, where $J = 17$ is the number of skeleton joints (corresponding to the shoulders, elbows, wrists, hips, knees, and ankles), and $D\in\left\{2, 3\right\}$ is spatial dimension of the coordinates. The underlying pose estimators---the fine-tuned domain-adapted infant pose (FiDIP) model \cite{huang2021invariant} for 2D and the heuristic weakly supervised 3D human pose estimation infant (HW-HuP-Infant) model \cite{liu2021heuristic} for 3D were specifically adapted for the infant domain.\\

\textbf{Infant-adult skeleton alignment--} One of the major challenges in the domain of skeleton-based action recognition lies in the significant variability of skeleton layouts across different datasets and scenarios. The diversity in joint definitions, proportions, scaling, and pose configurations across these layouts introduces complexity that directly impacts the efficacy of action recognition algorithms and makes transferring knowledge between two different datasets inefficient. The challenge of reconciling these layout differences and enabling robust recognition of actions regardless of skeletal variations is a critical concern in our studies. 
\skelayout

As shown in \autoref{fig:layout}, NTU RGB+D indicates the location of 25 joints in a 3D space. The layout of the infants 3D skeletons in the InfActPrimitive on the other hand, is based on the Human3.6M skeleton structure, which supports a total of 17 joints. To match the number of keypoints and align the skeleton data in these two datasets, We only select a subset of joins of NTU RGB+D skeleton that are common with the Human3.6M layout. We also reordered these joints, so the structures became as similar as possible. For the 2D skeletons, layouts of both NTU RGB+D and InfActPrimitive are based on the COCO structure.\\

\textbf{Action recognition--}
After preprocessing, we fed the extracted sequence of body keypoints from the input video into various state-of-the-art skeleton-based action recognition models leveraging different aspects of infant-specific pose representations. We categorize these skeleton-based models into three groups: CNN-based, graph-based, and RNN-based models to fully exploit the information encoded in the pose data and perform a comprehensive comparative analysis of the results.
\begin{itemize}


    \item{\textbf{Recurrent neural network structures}}
capture the long-term temporal correlation of spatial features in the skeleton.  We applied the part-aware LSTM (P-LSTM)\cite{shahroudy2016ntu} to segment body joints into five part groups and used independent streams of LSTMs to handle each part. At each timeframe $t$, the input $x^t$ is broken into $\left(x^t_1,\dots,x^t_P\right)$ parts, corresponding to $P$ parts of the body. These inputs are fed into $P$ streams of LSTM modules, where each LSTM has its own individual input, forget, and modulation gates. However, the output gate of these streams will be concatenated and will be shared among the body parts and their corresponding LSTM streams.

\item {\textbf{Graph convolutional networks (GCNs)}} represent skeletal data as a graph structure, with joints as nodes and connections as edges. To capture temporal relationships, we applied ST-GCN, which considers inter-frame connections between the same joints in consecutive frames. Furthermore, we employed InfoGCN \cite{chi2022infogcn}, which integrates a spatial attention mechanism to understand context-dependent joint topology, enhancing the existing skeleton structure. InfoGCN utilizes an encoder with graph convolutions and attention mechanisms to infer class-specific characteristics. $\mu_c$ and diagonal covariance matrix of a multivariate Gaussian distribution $\sigma_c$.  With an auxiliary independent random noise $\epsilon \sim N (0,I)$, $Z$ is sampled as $Z = \mu_c + \Sigma_c\epsilon$. The decoder block of the model, composed of a single linear layer and a softmax function, converts the latent vector $Z$ to the categorical distribution.


\item{\textbf{3D convolutional networks}} are mainly employed in RGB-based action recognition tasks to capture both spatial and temporal features across consecutive frames. To utilize the capabilities of a CNN-based framework, We first convert  keypoints in each frame into heatmaps. These heatmaps were generated by creating Gaussian maps centered at each joint within the frame. Subsequently, we applied the PoseC3D \cite{posec3d} method, which involved stacking these heatmaps along the temporal dimension, enabling 3D-CNNs to effectively handle skeleton-based action detection. Lastly, the representations extracted from each input sequence using the 3D convolutional layer were fed into a classifier. This classifier consists of a single linear layer followed by a softmax function, ultimately yielding the final class distribution.
\end{itemize}
\posresults

\section{Experimental Results}
In this section, we assess the performance of the models presented in our pipeline. We begin by providing an overview of our experimental setup and the datasets employed. Subsequently, we present the outcomes of various experiments. Finally, we conduct ablation studies and delve into potential avenues for future enhancements.

\subsection{Evaluation Datasets}



\textbf{NTU RGB+D}\cite{shahroudy2016ntu} is a large-scale action recognition dataset with both RGB frames and 3D skeletons. This dataset contains 56,000 samples across 60 action classes. Video samples have been captured by three Microsoft Kinect V2 camera sensors concurrently. 3D skeletal data contains the 3D locations of 25 major body joints at each frame. HRNet is used to estimate the 2D pose, which results in the coordination of 17 joints in the 2D space. Given that each video in this dataset features a minimum of two subjects, our approach involves evaluating the models within a cross-subject setting. In this particular setup, the models are trained using samples drawn from a designated subset of actors, while the subsequent evaluation is carried out on samples featuring actors who were not part of the training process. We have employed a train-test split paradigm that mirrors the methodology outlined in \cite{shahroudy2016ntu}. Specifically, we partition the initial cohort of 40 subjects into distinct training and testing groups, with each group composed of 20 subjects. In the context of this evaluative exercise, both the training and testing sets encompass a substantial number of samples, totaling 40, 320, and 16,560, respectively. It is noteworthy to mention that the training subjects for this particular evaluation bear the following identification numbers: 1, 2, 4, 5, 8, 9, 13, 14, 15, 16, 17, 18, 19, 25, 27, 28, 31, 34, 35, and 38. The remaining subjects have been thoughtfully reserved for the purpose of conducting rigorous testing.\\

\textbf{InfActPrimitive}, as detailed in \autoref{sec:infact}, combines video clips from two primary sources: data collected from the YouTube platform and data acquired through our independent data collection efforts. To evaluate our pipeline's performance on this dataset, the training set comprises all videos collected from YouTube, totaling 116 (sitting), 79 (standing), 62 (supine), 74 (prone), and 69 (all-fours) actions. Similarly, the test set consists exclusively of videos from our independently collected data, including 171 clips for sitting, 58 clips for standing, 62 clips for supine, 185 clips for prone, and 92 clips for all fours. This partitioning strategy enables us to assess the pipeline's ability to generalize across previously unobserved data and diverse sources, ensuring a comprehensive representation of various actions in both the training and test sets. This approach enhances the robustness of our evaluation by encompassing a wide range of settings and conditions found in YouTube videos and our collected data.

\subsection{Experimental Setup}
In this section, we detail the series of experiments conducted using our infant action recognition pipeline. We will also provide a comparative analysis, examining the outcomes in relation to the adult skeleton data.\\

\textbf{Baseline experiment--} In our baseline experiment, we trained various action recognition models, as detailed in \autoref{sec:pipeline}, separately on both the NTU RGB+D and InfActPrimitive datasets from scratch. With the exception of PoseC3D, all these models established baseline performance levels for both 2D and 3D-based action recognition tasks across both adult and infant domains. This baseline performance provides a starting point against which the performance of future experiments, such as fine-tuning or incorporating domain-specific knowledge, can be compared. We set the hyperparameter for ST-GCN, InfoGCN, deepLSTM and PoseC3D models a exactly as they were specified in \cite{stgcn},\cite{chi2022infogcn}, and \cite{posec3d}. In \autoref{tab:allResults}, the first pair of columns illustrate the experimental findings with 2D skeleton sequences from both the NTU RGB+D and InfActPrimitive datasets, respectively. Simultaneously, the fourth and fifth columns present the results in the context of 3D data. As demonstrated, PoseC3D consistently outperforms other models in both adult and infant action recognition domains. Nevertheless, a significant performance gap persists between infant and adult action recognition, which can be attributed to disparities in sample size and class distribution. The adult model benefits from a more abundant dataset, enabling it to effectively capture the spatiotemporal nuances of various actions, a characteristic that the InfActPrimitive dataset lacks. 
\confmatrixes

\autoref{fig:confs}
displays the confusion matrices for PoseC3D, InfoGCN, and ST-GCN methods. As illustrated, the sequences associated with the "Sitting" action class exhibit superior separability compared to other classes. However, it is evident that the InfoGCN model miserably fails in the infant action recognition
\tsne

\textbf{Transfer learning experiment--} To utilize the knowledge embedded in the adult action recognition, we initialized the model weights using the learned parameters obtained from prior training on the NTU RGB+D dataset. To address the substantial class disparities between the two datasets, we excluded the classifier weights, and for this experiment, initialized them randomly.

\

Given the significant disparity in the number of classes between the two datasets and the substantial impact of training set size on model performance, we chose to delve deeper into the implications of this experimental parameter. Notably, limited data availability posed challenges to achieving high accuracy in models trained on InfActPrimitive. To determine whether this issue extended beyond the domain of infant action recognition, we made modifications to the training subset of NTU RGB+D. Specifically, we curated a subset comprising only five action classes, namely, 'sit down,' 'stand up,' 'falling down,' 'jump on,' and 'drop,' which closely matched those in InfActPrimitive. We then restricted the number of samples per class in this subset to align with the size of the InfActPrimitive training subset. The validation samples for these selected classes remained unchanged.

As shown in Figure \ref{fig:tsne}, the latent variables demonstrate a significantly greater degree of separability within the adult domain compared to the infant domain. This finding highlights the potential limitations of models pretrained on infants in capturing the underlying patterns specific to the infant domain. The disparity can be attributed to the substantial differences between the adult and infant domains, emphasizing the necessity for domain-specific model adaptations or training approaches.




\textbf{Intra-class data diversity experiment--} In our final experiment, we investigate the impact of intra-class diversity on action recognition model performance. We hypothesize that the absence of structural coherence and the inherent variations among samples from the same class can significantly reduce validation accuracy. While traditional action recognition datasets like NTU RGB+D are known for rigid action instructions and minimal intra-class variation, our InfActPrimitive dataset, derived from in-the-wild videos, exhibits a higher level of variability in performed actions. To test this hypothesis, we conducted cross-validation training, dividing our training dataset into five subsets and training on four while validating on the fifth. The original validation set of InfActPrimitive was used for testing. Given the superior results achieved with the PoseC3D model using 2D skeleton data, we considered this model as an infant action recognition model. Our findings, presented in \autoref{tab:cross_val_results}, shed light on the influence of intra-class diversity on action recognition model performance.

\crossvalresults

As shown in \autoref{tab:cross_val_results}, although each experiment yields high training accuracy, there are substantial variations in validation and testing accuracies across experiments. These outcomes reveal discrepancies in the training datasets, leading to inconsistent learning, and underscore distinctions between videos collected from diverse sources.

\section{Conclusion}
Our work has introduced a unique dataset for infant action recognition, which we believe will serve as an invaluable benchmark for the field of infant action recognition and milestone tracking. Through our research, we applied state-of-the-art skeleton-based action recognition techniques, with Pose3D achieving reasonable performance. However, it is important to note that most other successful state-of-the-art action recognition methods failed miserably when it came to categorizing infant actions. This stark contrast underscores a significant knowledge gap between infant and adult action recognition modeling. This divergence arises from the distinct dynamics inherent in infant movements compared to those of adults, emphasizing the need for specialized, data-efficient models tailored explicitly for infant video datasets. Addressing this challenge is crucial to advancing the field of infant action recognition and ensuring that the developmental milestones of our youngest subjects are accurately tracked and understood. Our findings shed light on the unique intricacies of infant actions and pave the way for future research to bridge the gap in modeling techniques and foster a deeper understanding of infant development.

{\small
\bibliographystyle{ieee_fullname.bst}
\bibliography{refs}
}

\end{document}